\pdfoutput=1

\documentclass[11pt]{article}

\usepackage[final]{acl}

\usepackage{times}
\usepackage{latexsym}

\usepackage[T1]{fontenc}

\usepackage[utf8]{inputenc}

\usepackage{microtype}

\usepackage{graphicx}
\usepackage{booktabs}
\usepackage{pifont} 
\usepackage{amsfonts}

\usepackage{inconsolata}
\newcommand{\wt}{\textcolor[RGB]{0,0,0}}
\newcommand{\revise}{\textcolor[RGB]{0,0,0}}
\title{Fine-Grained Image-Text Alignment in Medical Imaging \\ Enables Explainable Cyclic Image-Report Generation}

\author{Wenting Chen$^{1}$ \quad Linlin Shen$^3$ \quad Jingyang Lin$^{4}$  \quad Jiebo Luo$^{4}$  \\  \bf Xiang Li$^{5}$$^*$ \quad \bf Yixuan Yuan$^2$\thanks{Xiang Li and Yixuan Yuan are corresponding authors.} \\
$^1$City University of Hong Kong \qquad $^2$The Chinese University of Hong Kong \\
$^3$Shenzhen University \quad \quad  $^4$University of Rochester \\
$^5$Massachusetts General Hospital and Harvard Medical School \\
$^1${wentichen7-c@my.cityu.edu.hk}   \quad $^2${yxyuan@ee.cuhk.edu.hk} \quad $^3${llshen@szu.edu.cn} \\ 
$^4${\{jluo@cs, jlin81@ur\}.rochester.edu}\quad $^5${xli60@mgh.harvard.edu} \\
}
\begin{document}
\maketitle
\begin{abstract}
Fine-grained vision-language models (VLM) have been widely used for inter-modality local alignment between the predefined fixed patches and textual words. However, in medical analysis, lesions exhibit varying sizes and positions, and using fixed patches may cause incomplete representations of lesions. 
Moreover, these methods provide explainability by using heatmaps to show the general image areas potentially associated with texts rather than specific regions, making their explanations not explicit and specific enough.
To address these issues, we propose a novel Adaptive patch-word Matching (AdaMatch) model to correlate chest X-ray (CXR) image regions with words in medical reports and apply it to CXR-report generation to provide explainability for the generation process. 
\wt{AdaMatch exploits the fine-grained relation between adaptive patches and words to provide explanations of specific image regions with corresponding words. To capture the abnormal regions of varying sizes and positions, we introduce an Adaptive Patch extraction (AdaPatch) module to acquire adaptive patches for these regions adaptively.}
Aiming to provide explicit explainability for the CXR-report generation task, we propose an AdaMatch-based bidirectional LLM for Cyclic CXR-report generation (AdaMatch-Cyclic). It employs AdaMatch to obtain the keywords for CXR images and `keypatches' for medical reports as hints to guide CXR-report generation. Extensive experiments on two publicly available CXR datasets validate the effectiveness of our method and its superior performance over existing methods. 
\end{abstract}
\section{Introduction}
\label{sec:intro}
\begin{figure}[t]
  \centering
   \includegraphics[width=\linewidth]{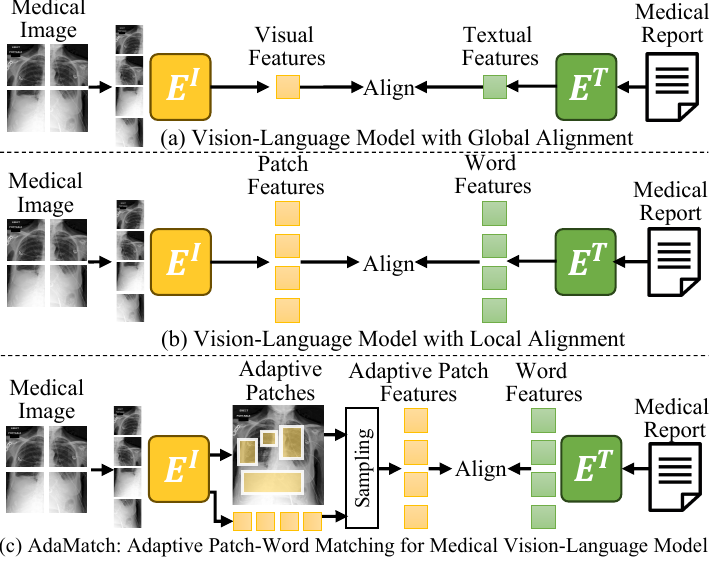}
   \caption{ Current vision-language models (VLM) achieve (a) global alignment and (b) local alignment by matching overall visual with textual features, and aligning patches with word features, respectively. (c) To exploit the relation between textual words and abnormal patches with varied sizes, our AdaMatch obtains adaptive patch features and aligns them with word features.}
   \label{fig:subview}
\end{figure}

Inter-modality alignment, such as vision and language, has been an important task with growing interests in the field of computer vision, especially with the recent advancement in representation learning~\cite{radford2021learning}. Technologies like contrastive learning and self-supervised learning have dramatically improved state-of-the-art alignment performance. Recent vision-language models (VLMs) demonstrate two approaches: global contrastive alignment, which integrates images and texts at a global level~\cite{radford2021learning,jia2021scaling,jang2023unifying,wang2023image,yang2022vision}, and local alignment, focusing on detailed connections between visual objects and textual words~\cite{chen2020uniter,li2020oscar,li2020unimo,zhan2021product1m,kim2021vilt,yao2021filip}, as illustrated in Fig.~\ref{fig:subview}.

Current VLMs with local alignment either adopt the pre-trained object detector to extract region-of-interest (ROI) features from images and match the corresponding object features with textual words~\cite{chen2020uniter,li2020oscar,li2020unimo,zhan2021product1m}, or align the visual token from each patch and the textual token into the same embedding space~\cite{kim2021vilt,yao2021filip,ji2021improving, wang2022multi}. The former highly relies on the quality of the object detector and its predefined classes, which is less generalizable to new domains. The latter family of methods learns the alignment in a more automatic and data-driven manner. However, most of these methods depend on a pre-defined patch size and positions (e.g., grids) across images. In the most challenging cases, such as the analysis of medical image, lesions can exhibit a wide range of shapes, sizes, and positions. A fixed partition of image patches can lead to incomplete or ambiguous representations of the key imaging abnormalities. Therefore, it is highly desirable to adaptively exploit the fine-grained relationship between image embeddings derived from a more flexible patching scheme and textual embeddings.

Another challenge in the current VLMs lies in their explainability: it is generally difficult to delineate the image-text relationship learned by the model, especially for the current medical VLMs. 
Current solutions to provide such explanations in medical VLMs leverage the attention maps from the intermediate layer to visualize the location of the abnormalities~\cite{moon2022multi, huang2021gloria, yan2022clinical}. Other methods ~\cite{wan2023med, chen2023knowledge, liu2023imitate} utilize network gradients such as Grad-CAM~\cite{selvaraju2017grad} to generate the heatmaps according to lesion types based on ground-truth reports. However, both maps can only show the general areas potentially associated with the corresponding text data rather than pinpointing a specific region. In addition, gradient-based methods need ground-truth reports, prohibiting them from functioning correctly beyond training data. It is thus highly necessary to develop a mechanism that could provide explicit and specific explanations of input image or text during inference time.

\begin{figure*}[t]
  \centering
   \includegraphics[width=\linewidth]{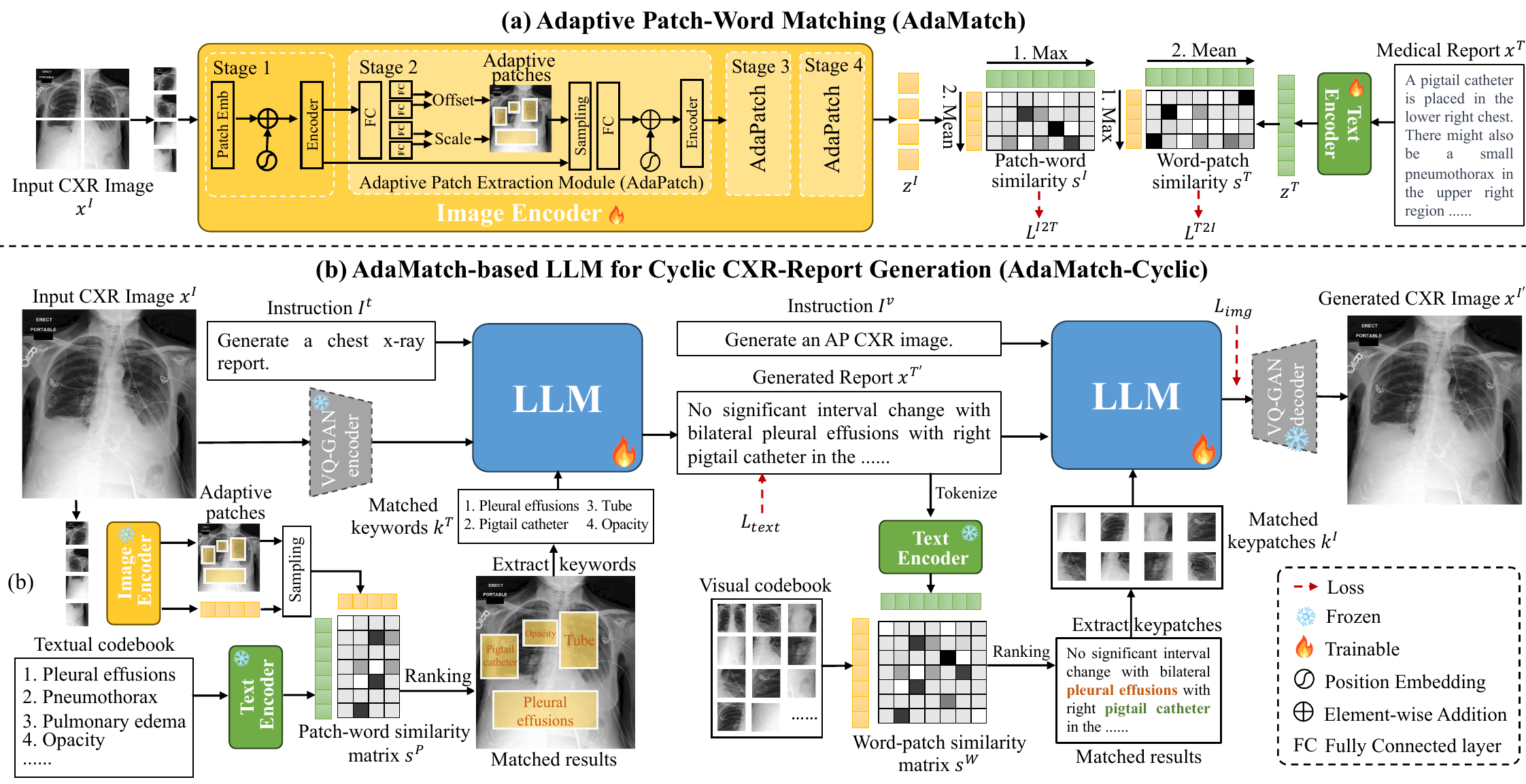}

   \caption{The overview of the proposed methods. (a) Adaptive patch-word Matching (AdaMatch) model. (b) AdaMatch-based bidirectional large language model (LLM) for cyclic CXR-report generation (AdaMatch-Cyclic).}
   \label{fig:overview}
\end{figure*}

To address these two challenges above, we propose a novel Adaptive patch-word Matching (AdaMatch) model to match fine-grained image regions of various sizes and positions with textual data. AdaMatch introduces an image encoder with multiple Adaptive Patch extraction (AdaPatch) modules to adaptively acquire the patches associated with certain text tokens. It then performs patch-word alignment based on contrastive learning. AdaMatch is specifically developed in the context of aligning radiology images (chest X-ray, CXR) and their corresponding radiology reports with the capability of achieving cyclic (CXR-to-report and report-to-CXR) generation based on the learned alignment. Our premise is that such a cyclic generation task would serve as the best use case and evaluation criterion for the desired fine-grained alignment. Also, fine-grained cyclic generation between CXR and report will provide natural explainability for how the model aligns two modalities: for any given text token, we can visualize its matching imaging manifestation; and for any image region within a CXR image, we can tell the type of lesion or anatomical region it belongs to.

To implement the cyclic CXR-report generation, we propose an AdaMatch-based bidirectional model (AdaMatch-Cyclic). AdaMatch-Cyclic employs AdaMatch to identify the keywords for CXR images and the `keypatches' for medical reports to guide the generation tasks. Since the potential keywords for CXR images cover a wide range and ground-truth reports cannot be used during inference, we predefine a textual codebook with the most common entities from medical reports as prior knowledge during fine-grained alignment. With the textual codebook, AdaMatch aligns it with the adaptive patches to obtain matched keywords to facilitate report generation. Next, a VQ-GAN model encodes the CXR image into image tokens, and a Large Language Model (LLM) takes image tokens, the matched keywords, and the instructions as input to generate medical reports. Similarly, we also build a visual codebook with the most commonly seen patches as `keypatches', and use AdaMatch to obtain the matched keypatches from given text reports as hints for CXR generation. Utilizing medical reports, matched keypatches, and instructions, LLM generates image tokens, subsequently decoded by the VQ-GAN model to produce the resulting CXR image. 
Our contributions are summarized as follows:

\begin{itemize}
\item To exploit the fine-grained relation between CXR image patches and words of medical reports, we propose an Adaptive patch-word Matching (AdaMatch) model to obtain adaptive patches for abnormal regions and perform alignment between them and texts in medical reports.

\item We devise an AdaMatch-based bidirectional LLM for Cyclic CXR-report generation (AdaMatch-Cyclic) to facilitate the bi-directional generation between CXR and reports. Moreover, we build the textual and visual codebook to utilize AdaMatch to extract useful keywords and keypatches for the report and CXR generation, respectively.
\item Experiments on two publicly available chest X-ray datasets demonstrate the effectiveness of our method and its superior performance over the state-of-art methods.
\end{itemize}
%

\section{Related Works}
\label{sec:related_works}

\subsection{Fine-Grained Vision-Language Models}
Recently, several fine-grained vision-language models (VLM) 
achieve the local alignment by exploiting the fine-grained relation between visual objects and textual words. Some methods~\cite{chen2020uniter,li2020oscar,li2020unimo,zhan2021product1m} employ the pre-trained object detector to obtain object features from images and align them with textual features, and others~\cite{kim2021vilt,yao2021filip, wang2022multi,ji2021improving,xu2023multi,jiang2023copa} aim to align the fixed patches with the textual words locally. 
The former relies on a precise object detector, and the latter focuses on the relation between fixed patches inside the predefined grid and words. 
However, varying lung lesion characteristics may cause these methods to divide them into separate patches, resulting in incomplete semantic information.
Thus, we devise an Adaptive patch-word Matching (AdaMatch) model to adaptively exploit the fine-grained relation between flexible patches and textual words.

\subsection{CXR-Report Generation}
Medical VLMs are widely used in downstream tasks for chest radiographs, including CXR-to-report generation~\cite{chen2020generating,chen2021cross,yang2021joint,wang2022cross,ITHN2023,yang2023radiology,shi2023granularity,huang2023kiut} and report-to-CXR generation~\cite{rombach2022high,chambon2022adapting,chambon2022roentgen,lee2023unified,lee2023llm,han2024advancing,shentu2024cxr,hou2023diversity,hashmi2024xreal,chen2024medical} tasks. 
For CXR-to-report generation, \citealp{chen2021cross} introduces a cross-modal memory network with shared memory to align images with texts to promote report generation performance.
In the {report-to-CXR generation} task, prior techniques create annotated CXR images from medical reports to augment training data and address privacy concerns, which are categorized into diffusion-based and transformer-based methods.
In this paper, we use the cyclic generation (i.e. CXR-to-report and report-to-CXR generation) as use case and evaluation criteria for our fine-grained alignment method (AdaMatch). We also design an AdaMatch-Cyclic to employ AdaMatch to improve the explainability of cyclic generation.

\section{Methods}
In Fig.~\ref{fig:overview}, we propose an adaptive patch-word matching (AdaMatch) model to associate CXR image regions with words in medical reports, and apply it to CXR-report generation to enhance explainability. 
Given an input CXR image $x^I$, the image encoder with several Adaptive Patch extraction modules (AdaPatch) predicts the location and scale of multiple adaptive patches and processes the adaptive patch embeddings $z^I$. With $z^I$ and text embeddings $z^T$ extracted by text encoder for the medical report $x^T$, we compute the similarities $s^I$, $s^T$ between adaptive patches and text tokens to calculate the contrastive loss $L_{I2T}, L_{T2I}$ to optimize AdaMatch. For CXR-to-report generation, we utilize the frozen AdaMatch model to match a predefined textual codebook with the input CXR image $x^I$ to obtain matched keywords $k^T$ and feed a LLM with $k^T$, the instruction, and image tokens encoded by VQ-GAN from $x^I$ to generate a medical report $x^{T'}$. Then, the AdaMatch model is used to match a predefined visual codebook with the generated report $x^{T'}$ to acquire the matched keypatches $k^I$. With instruction, $x^{T'}$ and $k^I$, LLM outputs image tokens of the generated CXR image $x^{I'}$, which is optimized through $L_{text}$ and $L_{img}$.

\subsection{Adaptive Patch-Word Matching (AdaMatch)}
Current VLMs~\cite{chen2020uniter,yao2021filip} achieve fine-grained alignment between visual objects and textual words, but they may split lung lesions into separate fixed patches due to various sizes of lung lesions. Thus, we propose an Adaptive patch-word Matching (AdaMatch) to locate the important regions, extract adaptive patches for these regions, and align them with corresponding textual words, as shown in Fig.~\ref{fig:overview} (a). 

\noindent\textbf{Image Encoder.}
To obtain the adaptive patches, the image encoder comprises four stages with feature maps of decreasing scales. Specifically, in the first stage, we first adopt the patch embedding module to split the CXR image $x^I\in \mathbb{R}^{H\times W\times C}$ into $N$ patches with fixed size $s\times s$ and project them to patch embeddings $z^{(i)}(1\leq i \leq N)$ through a fully connected (FC) layer $g$,
    $z^{(i)}=g([ a^{(i,1)}; \dots;  a^{(i,s\times s)}])$,
where $a^{(i,j)}$ indicates the image features for the pixel located at the $q^{(i,j)}$. $q^{(i,j)}$ indicates the coordinates for the image pixel. $[\cdot]$ represents the concatenation operation among the features. Afterward, we pass the patch embeddings $z^{(i)}$ with the position embeddings $e_{pos}$ into a transformer encoder $R$ to obtain the outputs, $z^{(i)}=R(z^{(i)} + e_{pos})$.
To locate the potential lung lesions, we devise an \textbf{Adaptive Patch Extraction module (AdaPatch)} for the rest of stages. 
In AdaPatch, patch embeddings $z^{(i)}$ are fed into an FC layer, followed by four separate FC layers $f_1,f_2,f_3,f_4$ to predict offsets $(\delta_x^{(i)}, \delta_y^{(i)})$ and patch sizes $(s_w^{(i)}, s_h^{(i)})$ for adaptive patches, with offsets indicating shifts to the center $(c^{(i)}_{x},c^{(i)}_{y})$ of fixed patches,
\begin{equation}
    \delta_x^{(i)}=\small{Tanh}(f_1(z^{(i)})),
    \delta_y^{(i)}=\small{Tanh}(f_2(z^{(i)})),
\end{equation}
\begin{equation}
    s_w^{(i)}=\small{ReLU}(\small{Tanh}(f_3(z^{(i)})),
    s_h^{(i)}=\small{ReLU}(\small{Tanh}(f_4(z^{(i)})).
\end{equation}
With the offset and patch size, we compute the position of left-top $(a^{(i)}_{x},a^{(i)}_{y})$ and right-bottom corners $(b^{(i)}_{x},b^{(i)}_{y})$ for each adaptive patch,
\begin{equation}
    a^{(i)}_{x}=c^{(i)}_{x} + \delta_x^{(i)} - \frac{s_w^{(i)}}{2},
    a^{(i)}_{y}=c^{(i)}_{y} + \delta_y^{(i)} - \frac{s_h^{(i)}}{2},
\end{equation}
\begin{equation}
    b^{(i)}_{x}=c^{(i)}_{x} + \delta_x^{(i)} + \frac{s_w^{(i)}}{2},
    b^{(i)}_{y}=c^{(i)}_{y} + \delta_y^{(i)} + \frac{s_h^{(i)}}{2},
\end{equation}
and uniformly sample $m\times m$ feature points inside the patches. Since the coordinates may be fractional, bilinear interpolation is used to obtain the sampled feature points $\left \{\hat{p}^{(j)}\right \}_{1\leq j\leq m\times m}$. The embeddings of all sampled points are flattened and fed into an FC layer $f_5$ to obtain patch embeddings, $z^{(i)}=f_5([\hat{p}^{(i,1)};\dots;\hat{p}^{(i,m\times m)}])$.
Finally, we pass $z^{(i)}$ with position embeddings $e_{pos}$ into a transformer encoder $R$.
The final adaptive patch embeddings $z^I\in\mathbb{R}^{N\times d}$ are the ensemble of $z^{(i)}(1 \leq i \leq N)$, 
    $z^I = E^I(x^I) = \left \{z^I_1,\cdots,z^I_N\right \},$
where $E^I$, $x^I$, and $d$ denote the image encoder, the CXR image, and the dimension of $z^I$, respectively. 

\noindent\textbf{Text Encoder.}
We adopt a pre-trained text encoder $E^T$ to encode the medical report $x^T$ into text embeddings $z^T$. To make $z^T$ with the same dimension as $z^I$, we feed an FC layer with $z^T$ to reduce its dimension, 
    $z^T = E^T(x^T)=\left \{z^T_1,\cdots,z^T_K\right \} (z^T\in \mathbb{R}^{K\times d})$,
where $K$ is the number of text tokens.

\noindent\textbf{Patch-Word Alignment.}
To exploit the relation between the adaptive patches and textual tokens, we perform fine-grained contrastive representation learning to achieve patch-word alignment. Concretely, for the $i$-th CXR image $x^I_i$ and $j$-th medical report $x^T_j$, we first compute the similarities between all the adaptive patch embeddings $z^I_n (1 \leq n \leq N)$ and all the text embeddings $z^T_k (1 \leq k \leq K)$, and use the largest similarity $\max_{(1\leq k \leq K)}{{(z^I_n)}^\top z^T_k}$ as the patch-word maximum similarity for $n$-th adaptive patch embedding. Then, the patch-word maximum similarities for all the adaptive patch embeddings are averaged as the similarity $s^I_{i,j}$ of the $i$-th CXR image to the $j$-th medical report,
    $s^I_{i,j} = \frac{1}{N} \displaystyle\sum_{n=1}^{N} (z^I_n)^\top z^T_{m_n^I}$,
where $m_n^I = \arg \max_{(1\leq k \leq K)}{{(z^I_n)}^\top z^T_k}$.
Similarly, the similarity of the $j$-th medical report to the $i$-th CXR image is defined as,
    $s^T_{i,j} = \frac{1}{K} \displaystyle\sum_{k=1}^{K} (z^I_{m_k^T})^\top z^T_k$,
where $m_k^T = \arg \max_{(1\leq n \leq N)}{{(z^I_n)}^\top z^T_k}$. We exclude the padded textual tokens when computing the similarity.

With the cross-modal similarities $s^I_{i,j}$ and $s^T_{i,j}$ for the $i$-th CXR image and $j$-th medical report, we compute the adaptive patch-word contrastive loss $L_i^I$ for $i$-th CXR image $x_i^I$,
\begin{equation}
    L_i^{I2T}(x_i^I,\left\{ x_j^T \right\}_{j=1}^b) = - \frac{1}{b}\log\frac{exp(s^I_{i,i}/\tau)}{\textstyle\sum_{j}{exp(s_{i,j}^I/\tau)}},
\end{equation}
where $b$,  $\tau$, and $\left \{x_i^I, x_i^T \right\}$ represent the batch size, temperature hyperparameter and the positive CXR-report pair, respectively. Similarly, the adaptive word-patch contrastive loss $L_i^T$ for $i$-th medical report $x_i^T$ is formulated as,
\begin{equation}
    L_i^{T2I}(x_i^T,\left\{ x_j^I \right\}_{j=1}^b) = - \frac{1}{b}\log\frac{exp(s^T_{i,i}/\tau)}{\textstyle\sum_{j}{exp(s_{j,i}^T/\tau)}}.
\end{equation}
The final contrastive loss for a mini-batch is calculated by,
    $L = \frac{1}{2}\sum_{i=1}^{b}{(L_i^{I2T}+L_i^{T2I})}$.
With $L$, AdaMatch learns to locate important patches with varied sizes and exploits the fine-grained relation.
\subsection{AdaMatch-based LLM for Cyclic CXR-Report Generation (AdaMatch-Cyclic)}
To provide explicit explainability for CXR-report generation task, we propose an AdaMatch-based bidirectional LLM for the cyclic CXR-report generation (AdaMatch-Cyclic) by locating the potential lesions in CXR images and visualizing the appearance of description in reports to guide the generation process, as depicted in Fig.~\ref{fig:overview} (b). 


\subsubsection{CXR-to-Report Generation}
In CXR-to-report generation, we use AdaMatch to match a predefined textual codebook with the CXR image, providing keywords to guide LLM.

\noindent\textbf{Building textual codebook.}
Specifically, we first use a pre-trained BioEN~\cite{raza2022large} to extract the related entities from medical reports in the training set, where the related entities are divided into four entity groups, i.e., biological structure, detailed description, disease disorder, and sign symptom. Next, we compute the frequency of each entity and pick top $\kappa_0$ entities for each entity group as keywords in the textual codebook. 

\noindent\textbf{Keywords Extraction.}
With the textual codebook, we employ the frozen AdaMatch model to match keywords from the textual codebook with the adaptive patches of CXR images, and obtain a patch-word similarity matrix $s^P \in \mathbb{R}^{M\times N}$ between the keyword tokens and adaptive patch embeddings, where $M$ and $N$ denote the number of keyword tokens and adaptive patches. To extract the most matched keywords for each adaptive patch, we rank the patch-word similarity along the dimension of keyword tokens, obtain the top $\kappa_1$ patch-word similarities for each adaptive patch, and extract corresponding keywords $k^T$. The matched keywords can explain potential lesions in each adaptive patch, to assist LLM in generating medical reports.

\begin{figure}[t]
  \centering
   \includegraphics[width=\linewidth]{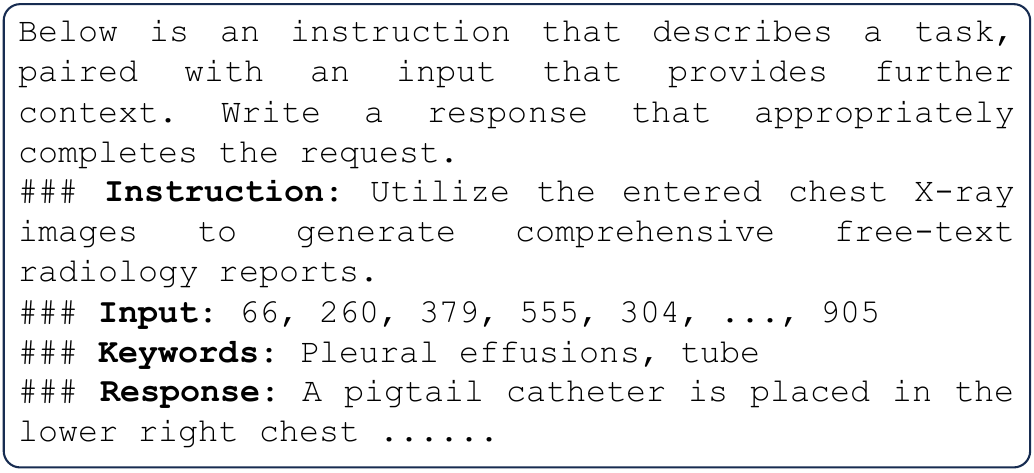}
   \caption{The example of instruction data for CXR-to-report generation.}
   \label{instruct_img2text}
\end{figure}

\noindent\textbf{Instruction tuning LLM.}
After keywords extraction, we use a frozen VQ-GAN~\cite{esser2021taming} encoder $E$ to encode the input CXR image $x^I$ to the quantized image latent vectors as image tokens $E(x^I)$. For CXR-to-report generation, we adopt the dolly-v2-3b~\cite{DatabricksBlog2023DollyV2} model as the pre-trained LLM, and convert CXR-to-report generation dataset in instruction-following format. 
An example of instruction data for CXR-to-report generation is depicted in Fig.~\ref{instruct_img2text}, with the instruction, input (the image tokens of the input CXR image), keywords (extracted by AdaMatch), and response parts (the ground-truth medical report). 
During training, the LLM learns to generate the hidden response part in an autoregressive manner.

By adopting AdaMatch-Cyclic, we can locate potential lesions and interpret them with keywords to enhance explainability. 


\subsubsection{Report-to-CXR Generation}
In report-to-CXR generation, AdaMatch matches a predefined visual codebook with generated reports $x^{T'}$, providing keypatches as guidance for the LLM to synthesize CXR images, where keypatches are important image patches related to the reports.

\noindent\textbf{Building visual codebook.} 
Concretely, we first construct a visual codebook with the most common adaptive patches as keypatches. To collect the most common adaptive patches for CXR images in the training set, AdaMatch matches the adaptive patches of CXR images with textual tokens of medical reports to obtain top $\kappa_2$ CXR-report pairs with the highest report-to-CXR similarities $s^T$. For each CXR-report pair, we compute the word-patch maximum similarity $\max_{(1\leq n \leq N)}{{(z^I_n)}^\top z^T_k}$ for each textual token, rank the word-patch maximum similarities, and extract the adaptive patches for top $\kappa_3$ similarities as keypatches in the visual codebook. Each keypatch includes its adaptive patch and the corresponding features.

\noindent\textbf{Keypatches Extraction.} 
With the visual codebook, the frozen AdaMatch matches the features of keypatches in the visual codebook with textual tokens of the generated report to acquire the word-patch similarity matrix $s^W \in \mathbb{R}^{(\kappa_2\times \kappa_3)\times K}$, where $K$ is the number of textual tokens. To obtain keypatches related to the generated report, we rank the word-patch similarity along the dimension of keypatches, obtain the top $\kappa_4$ word-patch similarity for each textual token, and extract the features of corresponding keypatches $k^I$. 

\begin{table*}[t]
\small
\setlength\tabcolsep{3.4pt}
	\renewcommand{\arraystretch}{1}
	\centering
	\caption{Comparison of CXR-to-report generation performance on the MIMIC-CXR and the OpenI datasets.}
 \begin{tabular}{c|cccccc|cccccc}
			\toprule[1pt] 
   & \multicolumn{6}{c|}{MIMIC-CXR} & \multicolumn{6}{c}{OpenI} \\ \hline
    Methods  & B-1 & B-2 & B-3 & B-4 & M & R-L & B-1 & B-2 & B-3 & B-4 & M & R-L \\ \hline
R2Gen & 0.3553  & 0.2232  & 0.1523  & 0.1038  & 0.1412  & 0.2784 & 0.3992 & 0.2407 & 0.1518 & 0.0973 & 0.1390 & 0.3052  \\ 
R2GenCMN & 0.3719  & 0.2332  & 0.1538  & 0.1053  & 0.1501  & 0.2827 & 0.4091 & 0.2493 & 0.1594 & 0.1045 & 0.1509 & 0.3181  \\
Joint-TriNet & 0.3585  & 0.2266  & \textbf{0.1550}  & 0.1021  & 0.1425  & 0.2788 & 0.3833 & 0.2409 & 0.1598 & 0.1078 & 0.1457 & 0.3293  \\
XProNet & 0.3532  & 0.2212  & 0.1498  & 0.1052  & 0.1415  & 0.2811 & 0.4114 & 0.2502 & 0.1598 & 0.1045 & 0.1457 & 0.3240  \\
ITHN & 0.3623 & 0.2128 & 0.1402 & 0.0992 & 0.1488 & 0.2622 & 0.2661 & 0.1516 & 0.0976 & 0.0663 & 0.1561 & 0.2617  \\
M2KT & 0.3661  & 0.2192  & 0.1465  & 0.1044  & 0.1528  & 0.2673 & 0.2559 & 0.1381 & 0.0819 & 0.0523 & 0.1468 & 0.2439  \\ \hline
AdaMatch-Cyclic & \textbf{0.3793}  & \textbf{0.2346}  & 0.1540  & \textbf{0.1060}  & \textbf{0.1625}  & \textbf{0.2859} & \textbf{0.4161} & \textbf{0.3002} & \textbf{0.2073} & \textbf{0.1446} & \textbf{0.1621} & \textbf{0.3656} \\
			\bottomrule[1pt]
	\end{tabular}
	\label{tab:report_gen}
\end{table*}

\begin{figure}[t]
  \centering
   \includegraphics[width=\linewidth]{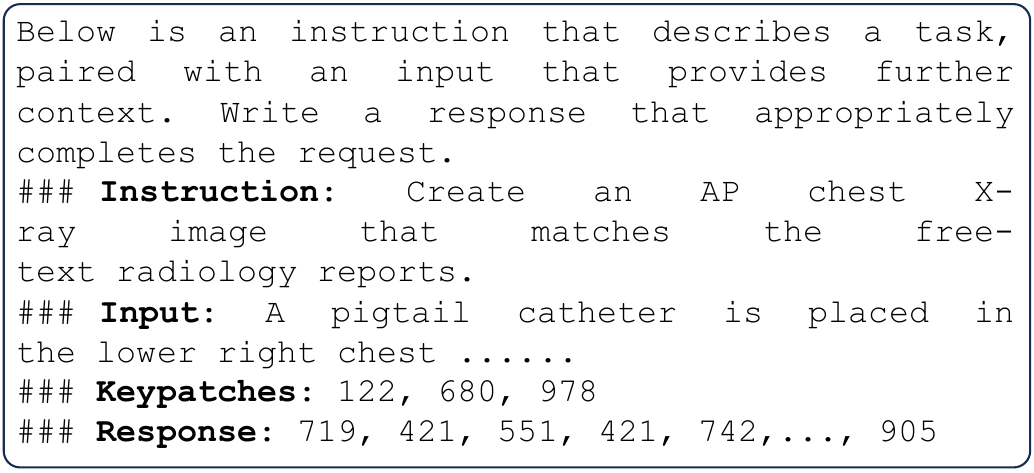}
   \caption{The example of instruction data for report-to-CXR generation.}
   \label{instruct_text2img}
\end{figure}

\noindent\textbf{Instruction tuning LLM.} 
After keypatches extraction, we use a frozen VQ-GAN encoder to convert the matched keypatches $k^I$ into image tokens $E(k^I)$, and feed the LLM with the instruction, generated report, and image tokens of keypatches in the instruction-following format, as shown in Fig.~\ref{instruct_text2img}. Then, LLM predicts image tokens, decoded by the VQ-GAN into the generated CXR image $x^{I'}$.

AdaMatch-Cyclic allows us to interpret generated reports with matched keypatches, thereby providing explainability to the generation procedure.
\subsubsection{Overall Objective}
To optimize AdaMatch-Cyclic, we use the standard language modeling objective for both CXR-to-report and report-to-CXR generation. For CXR-to-report generation, LLM receives the instruction $I^t$, image tokens of input CXR image $E(x^I)$, and matched keywords $k^T$, aiming to generate the ground-truth medical report of response part autoregressively. 
We compute the conditional probability of the $k$-th token,
    $P(u_k) = \mathcal{LLM}(I^t, E(x^I), k^T)$,
where $k$ is the token index after the response key (\texttt{\#\#\# Response:}).
The report generation loss for the response area is calculated by, 
    $L_{text} = \sum_{i=k}^{n} - \log P(u_i|u_1,u_2,\cdots,u_{i-1})$,
where $[u_1,u_2,\cdots,u_{i-1}]$ denotes the tokenized texts before the response part, and $n$ represents the maximum length of output tokens.
Similarly, for report-to-CXR generation, we feed the LLM with the instruction $I^v$, generated report $x^{T'}$, and image tokens of keypatches $E(k^I)$, and obtain the conditional probability of $k$-th token $w_k$, $P(w_k) = \mathcal{LLM}(I^v, x^{T'}, E(k^I))$. The CXR generation loss is defined as: 
   $L_{img} = \sum_{i=k}^{n} - \log P(w_i|w_1,w_2,\cdots,w_{i-1})$.
With $L_{text}$ and $L_{img}$, LLM can implement bidirectional CXR and report generation according to instructions.

\begin{figure*}[t]
  \centering
   \includegraphics[width=\linewidth]{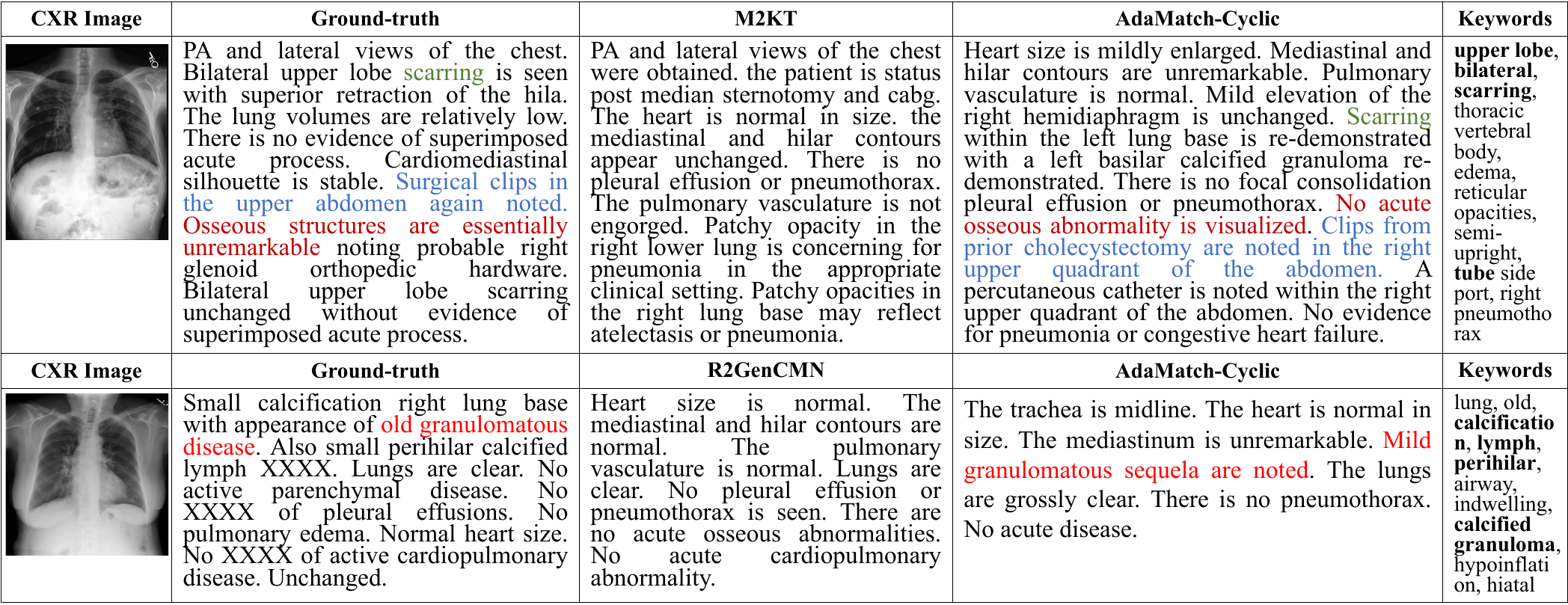}
   \caption{
Qualitative comparison of CXR-to-report generation on the MIMIC-CXR (1st row) and the OpenI (2nd row) datasets, highlighting similar meanings in colored text. The keywords are obtained from AdaMatch.}
   \label{fig:example_cxr2report}
\end{figure*}

\section{Experiments}
\subsection{Experiment Setting}
\noindent\textbf{Datasets.} 
We experiment on two main publicly available chest X-ray datasets, i.e. MIMIC-CXR~\cite{johnson2019mimic} and OpenI~\cite{demner2016preparing} datasets.
\textbf{MIMIC-CXR} dataset comprises 473,057 images and 206,563 reports from 63,478 patients. We use the official splits, i.e. 368,960 for training, 2,991 for validation, and 5,159 for testing. \textbf{OpenI} contains 3,684 report-image pairs with 2,912 for training and 772 for testing. Unlike prior methods, we utilize both finding and impression sections as medical reports. 

\begin{table}[t]
\small
\setlength\tabcolsep{5pt}
	\renewcommand{\arraystretch}{1.2}
	\centering
	\caption{
Performance of CXR-to-report generation compared to GPT-4V on 50 selected MIMIC-CXR cases.}
 \begin{tabular}{c|cccc}
			\toprule[1pt] 
    Methods  & B-1 & B-2 & B-3 & B-4 \\ \hline
R2Gen & 0.3513 & 0.2174 & 0.1447 & 0.1025 \\
R2GenCMN & 0.3587 & 0.2200 & 0.1436 & 0.0969 \\
Joint-TriNet &  0.3596 & 0.2218 & 0.1481 & 0.1026 \\
XProNet & 0.3356 & 0.2071 & 0.1374 & 0.0941 \\
ITHN &   0.3301 & 0.1839 & 0.1121 & 0.0723 \\
M2KT & 0.3626 & 0.2123 & 0.1391 & 0.0957 \\
GPT-4V & 0.2275 & 0.0878 & 0.0378 & 0.0166 \\ \hline
AdaMatch-Cyclic & \textbf{0.3754} & \textbf{0.2303} & \textbf{0.1520} & \textbf{0.1058} \\
			\bottomrule[1pt]
	\end{tabular}
	\label{tab:report_gen_selected}
\end{table}

\begin{table}[t]
\small
	\renewcommand{\arraystretch}{1.1}
	\centering
	\caption{Comparison of report-to-CXR generation performance on the MIMIC-CXR and the OpenI datasets.}
	\scalebox{0.95}{
 \begin{tabular}{c|c|c|c|c}
			\toprule[1pt] 
   
 & \multicolumn{2}{c|}{MIMIC-CXR} & \multicolumn{2}{c}{OpenI} \\ \hline
Methods & FID$\downarrow$ & NIQE$\downarrow$ & FID$\downarrow$ & NIQE$\downarrow$  \\ \hline
Stable diffusion & 9.2334  & 3.7894 & 8.2946  & 6.3496 \\ 
Adapting-Med &  8.2758 & 3.8871 &	5.8557  & 4.6534  \\ 
RoentGen & 9.5411 & 3.8834 & 6.5675 & 4.9085 \\ 
UniXGen & 6.7212 & 3.7125  & 11.9890 & 4.6610  \\ 
LLM-CXR & 2.1788  & 3.5969  & 1.6597 & 3.8206  \\ \hline
AdaMatch-Cyclic & \textbf{1.0916} & \textbf{3.3931}  & \textbf{1.5938} & \textbf{3.3096} \\ 
			\bottomrule[1pt]
	\end{tabular}
	}
	\label{tab:fid}
\end{table}

\noindent\textbf{Implementation Details.} 
In the AdaMatch-Cyclic model, we first train the AdaMatch model and then use the frozen AdaMatch to train LLM.
We adopt VQ-GAN~\cite{esser2021taming} models pre-trained on the MIMIC-CXR and OpenI datasets, respectively, and the dolly-v2-3b~\cite{DatabricksBlog2023DollyV2} model as pre-trained LLM. 
The source code will be released.
Please see appendix~\ref{appendix:implement} for more details. 

\noindent\textbf{Evaluation Metrics.}
We assess CXR-to-report generation using BLEU (B), METEOR (M), and ROUGE-L (R-L)~\cite{chen2020generating}, and report-to-CXR generation using FID~\cite{heusel2017gans} and NIQE~\cite{mittal2012making}.
The retrieval performance between CXR and report is assessed through the exact report in the top K retrieved reports for a given CXR image (R@K, K=$\left \{ 1,5,10\right \}$). 
 
\subsection{Comparison with State-of-the-Arts.}

\subsubsection{CXR-to-Report Generation}
We compare AdaMatch-Cyclic with current CXR-to-report generation methods on the MIMIC-CXR and the OpenI datasets, including R2Gen~\cite{chen2020generating}, R2GenCMN~\cite{chen2021cross}, Joint-TriNet~\cite{yang2021joint}, XProNet~\cite{wang2022cross}, ITHN~\cite{ITHN2023}, and M2KT~\cite{yang2023radiology}.
Since these methods are mainly trained on the finding section, we reimplement them on both the finding and impression sections. 
In Table~\ref{tab:report_gen}, AdaMatch-Cyclic achieves the best performance of 0.3793 in BLEU-1 on the MIMIC-CXR dataset, with superior generalization to OpenI dataset using the same model.
Compared to GPT-4V, AdaMatch-Cyclic significantly outperforms in BLEU-1 by 0.1479, as listed in Table~\ref{tab:report_gen_selected}. 
\begin{figure}[t]
  \centering
   \includegraphics[width=\linewidth]{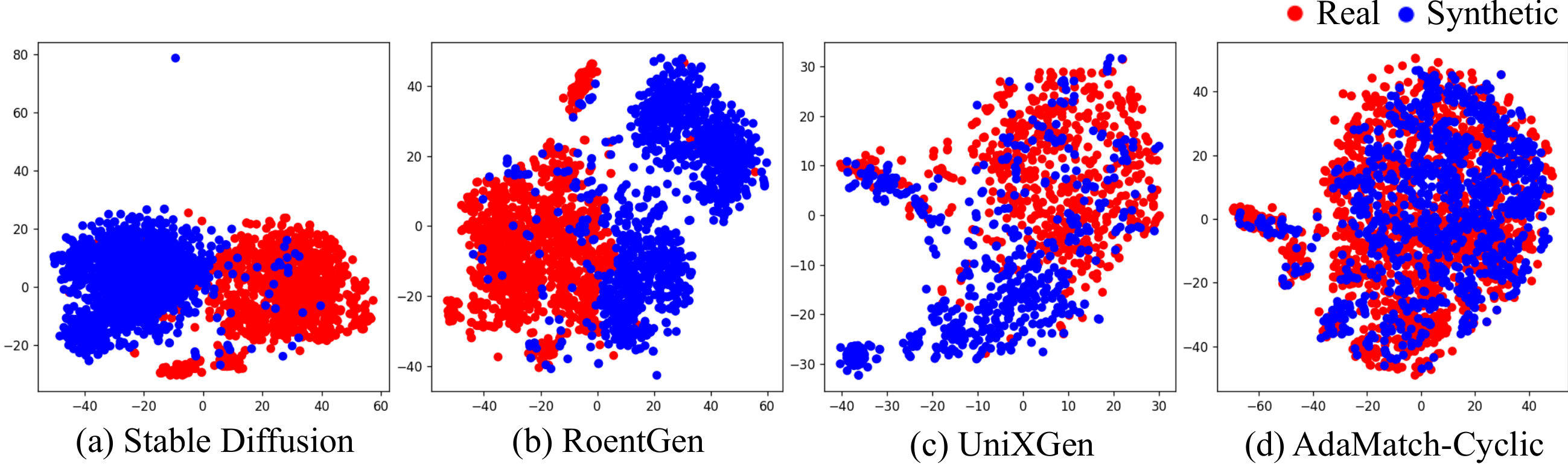}
   \caption{The t-SNE visualization of the real and synthetic CXR images on the MIMIC-CXR dataset.}
   \label{fig:tsne}
\end{figure}

\begin{figure}[t]
  \centering
   \includegraphics[width=\linewidth]{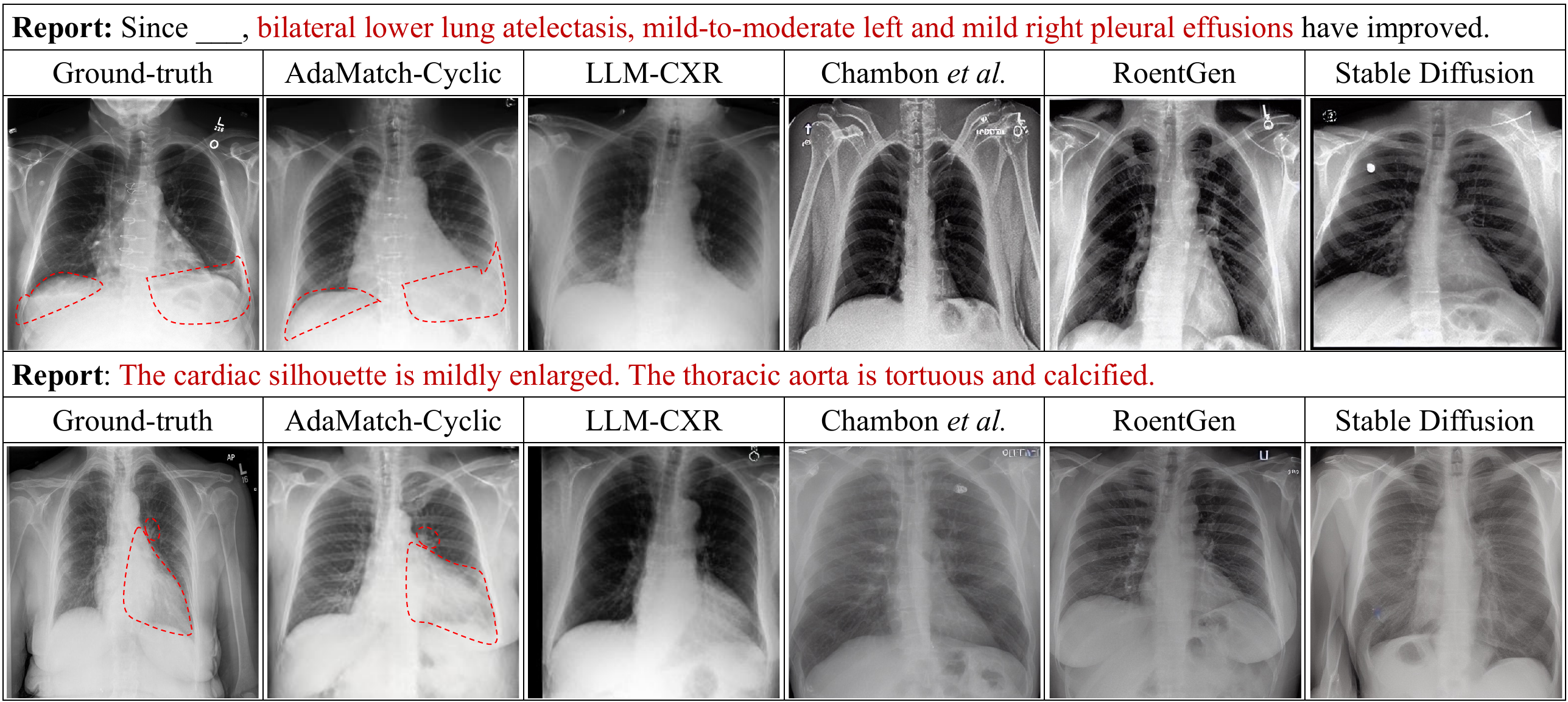}
   \caption{Generated CXR images of the MIMIC-CXR (1st row) and OpenI (2nd row) datasets with highlighted regions.}
   \label{fig:cxr_gen}
\end{figure}
\begin{table}[t]
\small
\setlength\tabcolsep{1pt} 
	\renewcommand{\arraystretch}{1.2}
	\centering
	\caption{Comparison of CXR-report retrieval performance (\%) on MIMIC-CXR dataset.}
 \scalebox{0.81}{
 \begin{tabular}{c|ccc|ccc}
			\toprule[1pt] 
   & \multicolumn{3}{c|}{CXR-to-Report} & \multicolumn{3}{c}{Report-to-CXR}  \\ \hline
Methods & R@1 & R@5 & R@10 & R@1 & R@5 & R@10 \\ \hline
\citet{fang2015captions} & 18.60 & 43.10 & 56.10 & 18.13 & 43.20 & 55.97 \\
\citet{chauhan2020joint} & 5.37 & 19.43 & 30.73 & 5.40 & 20.23 & 30.23 \\
ConVIRT~\cite{zhang2022contrastive} & 30.10 & 53.90 & 63.80 & 29.20 & 54.70 & 64.40 \\
GLoRIA~\cite{huang2021gloria} & 30.30 & 57.50 & 66.50 & 24.00 & 51.80 & 62.80 \\
JoImTeR-Net~\cite{ji2021improving} & 18.93 & 46.20 & 58.67 & 19.07 & 45.27 & 58.50 \\
MGCA~\cite{wang2022multi} & 25.80 & 51.90 & 62.10 & 27.90 & 51.20 & 61.60 \\
LIMITR~\cite{dawidowicz2023limitr} & 39.70 & 63.20 & 71.70 & 37.70 & 62.10 & 71.30 \\
Motor~\cite{lin2023towards} & 10.96 & 31.93 & 42.90 & 12.00 & 33.10 & 44.32 \\ \hline
AdaMatch & \textbf{51.47} & \textbf{86.19} & \textbf{94.77} & \textbf{51.18} & \textbf{86.46} & \textbf{94.60} \\
\bottomrule[1pt]
	\end{tabular}
    }
	\label{tab:retrieval}
\end{table}

\begin{table}[t]
\small
\setlength\tabcolsep{4.5pt}
	\renewcommand{\arraystretch}{1.1}
	\centering
	\caption{Ablation study on AdaMatch w/o AdaPatch.}
 \scalebox{0.93}{
 \begin{tabular}{c|ccc|ccc}
			\toprule[1pt] 
   & \multicolumn{3}{c|}{CXR-to-Report} & \multicolumn{3}{c}{Report-to-CXR}  \\ \hline
AdaPatch & R@1 & R@5 & R@10 & R@1 & R@5 & R@10 \\ \hline
\ding{55} & 48.77 & 83.89 & 92.94 & 48.72 & 83.95 & 92.90 \\
\ding{51} & \textbf{51.47} & \textbf{86.19} & \textbf{94.77} & \textbf{51.18} & \textbf{86.46} & \textbf{94.60} \\
\bottomrule[1pt]
	\end{tabular}
 }
	\label{tab:ablation_adapatch}
\end{table}

\begin{table*}[t]
\small
\setlength\tabcolsep{5pt}
	\renewcommand{\arraystretch}{1}
	\centering
	\caption{\revise{Effectiveness of report-to-CXR and CXR-to-report generation tasks.}}
 \scalebox{0.96}{
 \begin{tabular}{cc|cccccc|cc}
			\toprule[1pt] 
   Report-to-CXR & CXR-to-Report & B-1$\uparrow$ & B-2$\uparrow$ & B-3$\uparrow$ & B-4$\uparrow$ & M$\uparrow$ & R-L$\uparrow$ & FID$\downarrow$ & NIQE$\downarrow$ \\ \hline
 & \checkmark & 0.3542 & 0.1973 & 0.1190 & 0.0758 & 0.1329 & 0.2392 & - & - \\ 
\checkmark & & - & - & - & - & - & - & 1.7128 & 4.0391 \\ 
\checkmark & \checkmark & 0.3793 & 0.2346 & 0.1540 & 0.1060 & 0.1625 & 0.2813 & 1.0916 & 3.3931 \\ 
\bottomrule[1pt]
	\end{tabular}
 }
	\label{tab:cyclic}
\end{table*}

\begin{table}[t]
\small
\setlength\tabcolsep{4.5pt}
	\renewcommand{\arraystretch}{1.1}
	\centering
	\caption{\revise{Comparison of AdaPatch and other adaptive vision models.}}
 \scalebox{0.96}{
 \begin{tabular}{c|ccc|ccc}
			\toprule[1pt] 
   & \multicolumn{3}{c|}{CXR-to-Report} & \multicolumn{3}{c}{Report-to-CXR}  \\ \hline
Models & R@1 & R@5 & R@10 & R@1 & R@5 & R@10 \\ \hline
A-ViT & 49.78 & 83.14 & 92.77 & 49.49 & 83.19 & 92.94 \\
AdaViT & 50.14 & 84.52 & 93.20 & 50.54 & 84.10 & 93.83 \\
AdaPatch & \textbf{51.47} & \textbf{86.19} & \textbf{94.77} & \textbf{51.18} & \textbf{86.46} & \textbf{94.60} \\
\bottomrule[1pt]
	\end{tabular}
 }
	\label{tab:ablation_otherAda}
\end{table}

\begin{table}[t]
\small
\setlength\tabcolsep{5pt}
	\renewcommand{\arraystretch}{1.1}
	\centering
	\caption{CXR-report retrieval performance of AdaMatch with different stages.}
 \begin{tabular}{c|ccc|ccc}
			\toprule[1pt] 
   & \multicolumn{3}{c|}{CXR-to-Report} & \multicolumn{3}{c}{Report-to-CXR}  \\ \hline
Stage & R@1 & R@5 & R@10 & R@1 & R@5 & R@10 \\ \hline
2 & 42.28 & 78.75 & 89.85 & 42.81 & 78.96 & 89.67 \\
3 & \textbf{51.47} & \textbf{86.19} & \textbf{94.77} & \textbf{51.18} & \textbf{86.46} & \textbf{94.60} \\
4 & 49.68 & 83.04 & 92.67 & 49.39 & 83.09 & 92.84 \\
\bottomrule[1pt]
	\end{tabular}
	\label{tab:stages}
\end{table}

\begin{figure}[tp]
  \centering
   \includegraphics[width=\linewidth]{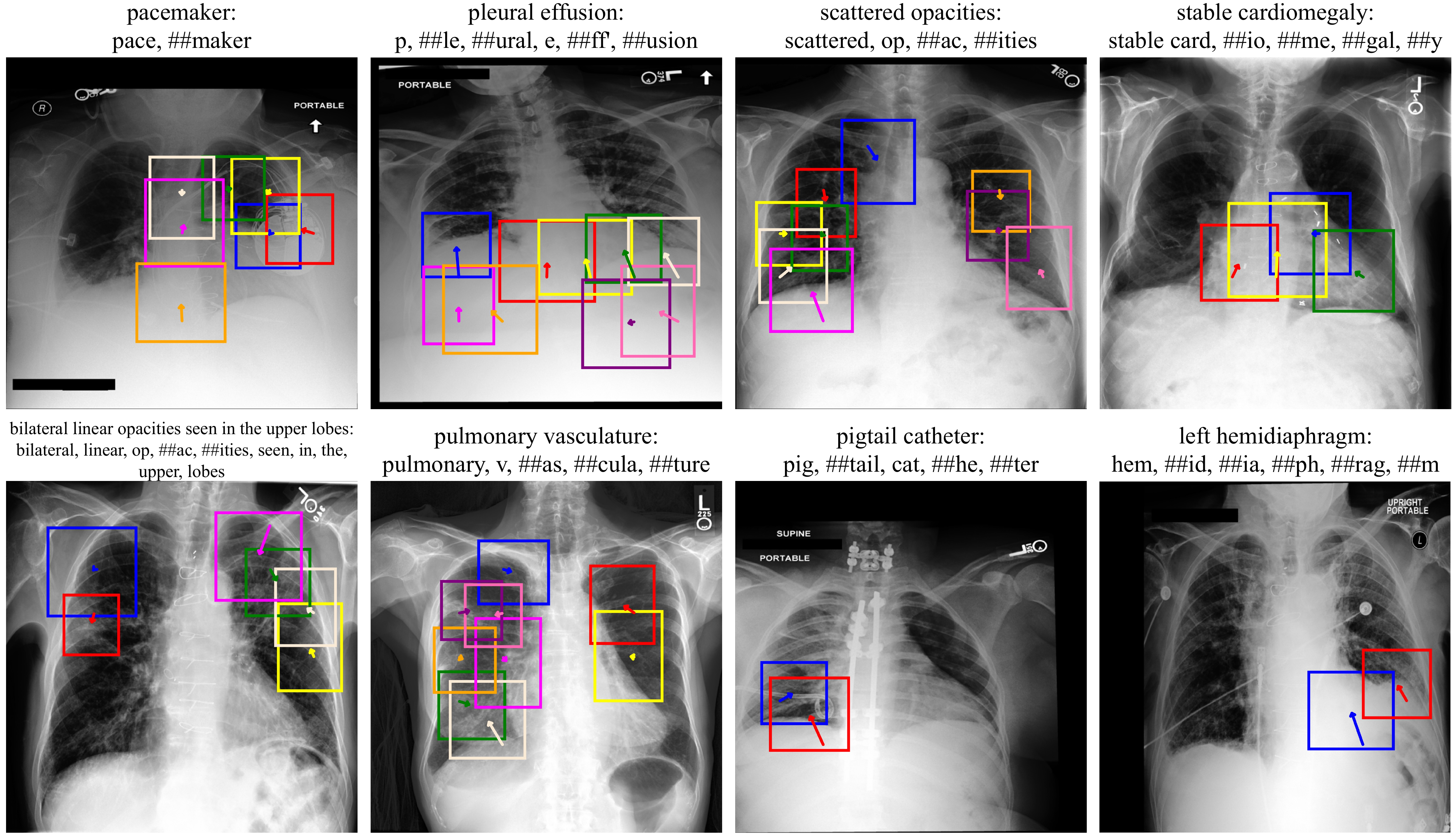} 
   \caption{\revise{Visualization of texts and its adaptive patches. Boxes and arrows in different colors show adaptive patches and their center shifts from fixed patches.}}
   \label{fig:appendix_vis_boxes}
\end{figure}

In Fig.~\ref{fig:example_cxr2report}, we compare AdaMatch-Cyclic's performance with M2KT and R2GenCMN. AdaMatch-Cyclic accurately captures relevant keywords like `scarring', and describes `osseous abnormality' and `clips in the right upper quadrant of the abdomen' in the first case, showcasing superior report quality over previous methods.

\subsubsection{Report-to-CXR Generation}
We quantitatively compare our method with text-to-image generation method like Stable Diffusion~\cite{rombach2022high} and report-to-CXR generation methods, such as Adapting-Med~\cite{chambon2022adapting}, RoentGen~\cite{chambon2022roentgen}, UniXGen~\cite{lee2023unified}, and LLM-CXR~\cite{lee2023llm}. In Table~\ref{tab:fid}, AdaMatch-Cyclic achieves the highest FID scores on both datasets, showing its superior effectiveness in generating CXR. To compare the high-level feature distribution of CXR images generated by different methods, we randomly select 1,000 cases from test set, and apply t-SNE visualization to the real and synthetic CXR images on the MIMIC-CXR dataset. In Fig.~\ref{fig:tsne}, while existing methods' synthetic CXR images differ from real ones, AdaMatch-Cyclic's almost overlap with real ones, indicating its superiority in report-to-CXR generation.
In Fig.~\ref{fig:cxr_gen}, we display generated CXR images from the MIMIC-CXR and OpenI datasets. In the first example, AdaMatch-Cyclic accurately synthesizes `left and right pleural effusions', while other methods fail to generate such features. This suggests the superior ability of our method to produce realistic CXR images based on input reports.


\subsubsection{CXR-Report Retrieval}
To demonstrate the superiority of AdaMatch, we evaluate the CXR-to-report and report-to-CXR retrieval performance on the MIMIC-CXR dataset compared to existing methods. We utilize the AdaMatch to compute the similarities between CXR images and reports, rank the similarities, and obtain the retrieval results. 
As shown in Table~\ref{tab:retrieval}, AdaMatch significantly outperforms LIMITR with R@1 of 11.77\% and 13.48\% for CXR-to-report and report-to-CXR retrieval. This indicates the superior ability of our method to extract distinct semantic features and align them accurately between CXR images and medical reports.

In Table 5, to ablate AdaPatch, we remove the layers to predict offset and scale. We then aligned grid features from the last stage of the pyramid vision Transformer with textual tokens to compute the image-text similarity for retrieval tasks.

\subsection{Ablation Study}

\noindent\textbf{Effectiveness of AdaPatch.}
We evaluate the effectiveness of AdaPatch by comparing the CXR-report retrieval performance of AdaMatch with and without AdaPatch. \revise{To ablate AdaPatch, we remove the layers to predict offset and scale, and compute the similarity between grid features of the last stage and textual tokens for retrieval tasks.} In Table~\ref{tab:ablation_adapatch}, AdaPatch significantly improves R@1 by approximately 3\% for both retrieval directions, highlighting its effectiveness. \revise{To demonstrate the optimality of the AdaPatch design, we compare AdaPatch with other adaptive vision models, such as A-ViT~\cite{yin2022vit} and AdaViT~\cite{meng2022adavit}, in the CXR-report retrieval task. As shown in Table~\ref{tab:ablation_otherAda}, AdaPatch exhibits superior retrieval performance compared with other methods.}

\noindent\textbf{\revise{Effectiveness of Cyclic Generation.}}
\revise{To validate the necessity of report-to-CXR and CXR-to-report generation tasks, we remove each task and evaluate their performance, as listed in Table~\ref{tab:cyclic}. When ablating report-to-CXR generation task, the performance of CXR-to-report generation decreases substantially compared to our method, and vice versa.
These suggest that the CXR-to-report and report-to-CXR generation can indeed benefit each other, and their interaction is crucial for the overall performance of the model.}


\noindent\textbf{Image Encoder with Different Stages.}
In AdaMatch, the image encoder comprises several stages, with AdaPatch modules in the stages. To assess the effectiveness of the stage number, we compare the retrieval performance of AdaMatch with two, three, or four stages. Table~\ref{tab:stages} shows that the three-stage image encoder achieves the highest R@1 of 51.47\%, indicating that the features from the 3rd stage are most beneficial for CXR-report retrieval.

\noindent\textbf{Visualization of Adaptive Patches.}
We visualize adaptive patches in AdaPatch of some examples from the MIMIC-CXR dataset in Fig.~\ref{fig:appendix_vis_boxes}. Each example includes text, its tokens, and the corresponding CXR image with marked adaptive patches. The patches are represented by colored bounding boxes with arrows indicating the center shift from the fixed patch. In the first example, adaptive patches cover the pacemaker and wire, and the second one can find pleural effusions, implying the capacity of AdaPatch to localize regions relevant to input text tokens accurately.

\section{Conclusion}
We propose AdaMatch for fine-grained image-text alignment, which presents the first work to adaptively associate image patches with words and improve the explainability of cyclic CXR-report generation. It includes AdaPatch to acquire adaptive patches for abnormal regions and performs patch-word alignment between adaptive patches with textual tokens. We implement cyclic CXR-report generation by using AdaMatch to provide explanations for the generation process. In addition, the fine-grained cyclic generation process provides a natural explainability for the alignment between CXR images and reports. Extensive experiments on two CXR datasets show the effectiveness of our method and its superiority over previous methods.


\section*{Limitation}
While our research has made significant strides in utilizing chest X-ray datasets, it is important to acknowledge certain limitations. Our experiments predominantly focus on chest X-ray datasets due to their availability of large-scale images paired with high-quality medical reports. However, these datasets primarily consist of patients from the ICU, potentially skewing our model towards severe disease domains. In our future endeavors, we intend to expand the scope of our methodology by applying the proposed AdaMatch-Cyclic to multi-domain scenarios, thereby mitigating this limitation and enhancing the versatility of our approach.

\section*{Acknowledge}
This work was supported by the Hong Kong Research Grants Council (RGC) General Research Fund under Grant 14220622, Innovation and Technology Commission Innovation and Technology Fund ITS/229/22, and the National Natural Science Foundation of China under Grant 82261138629.

\bibliography{custom}
\clearpage
\appendix

\noindent { \LARGE \textbf{Appendix}}

\smallskip
\smallskip
\smallskip

\noindent\textbf{Abstract.} In this supplementary material, we provide additional information about the proposed method.
Appendix~\ref{appendix:ablation} illustrates ablation studies on AdaMatch-Cyclic. Appendix~\ref{sec:vis} demonstrates visual results of CXR-to-report generation, report-to-CXR generation, and adaptive patches. Appendix~\ref{appendix:implement} provides the implementation details of the proposed method.

\section{Ablation Studies on AdaMatch-Cyclic}
\label{appendix:ablation}
In the proposed AdaMatch-Cyclic, we leverage the AdaMatch to obtain keywords for CXR-to-report generation and keypatches for the report-to-CXR generation. To analyze the effectiveness of keywords and keypatches for the generation process, we ablate them in the AdaMatch-Cyclic on the MIMIC-CXR dataset to evaluate the CXR-report generation performance. \\

\noindent\textbf{Effectiveness of keywords.}
As listed in Table~\ref{tab:ablation}, we ablate both keywords and keypatches of AdaMatch-Cyclic to create a baseline model. When we employ the keywords in the baseline model, the CXR-to-report generation performance improves significantly by about 0.03 in BLEU-4, indicating the effectiveness of the keywords obtained from AdaMatch. To further analyze the influence of the number of keywords $N_w$, we train the proposed AdaMatch-Cyclic with different $N_w=\left \{10,15,20\right \}$. In Table~\ref{tab:num_keywords}, AdaMatch-Cyclic achieves the best report generation performance with the ROUGE-L of 0.2859, when the $N_w$ is set to 10.\\
\begin{table}[b]
\small
\setlength\tabcolsep{3pt}
	\renewcommand{\arraystretch}{1.1}
	\centering
	\caption{The effectiveness of keywords and keypatches.  B-4, M,  and R-L represent BLEU-4, METEOR, and ROUGE-L, respectively.}
 \begin{tabular}{cc|cccc}
			\toprule[1pt] 
    Keywords & Keypatches  & B-4 & M & R-L & FID \\ \hline
 &  &  0.0778  & 0.1295  & 0.2402  & 1.5378  \\ \hline
\checkmark &    & 0.1059  & 0.1649  & 0.2813  & 1.5132  \\ \hline
\checkmark & \checkmark &  0.1060  & 0.1625  & 0.2859  & 1.0916  \\ 
			\bottomrule[1pt]
	\end{tabular}
	\label{tab:ablation}
\end{table}

\begin{table}[t]
\small
\setlength\tabcolsep{3.5pt}
	\renewcommand{\arraystretch}{1.2}
	\centering
	\caption{The analysis on the different number of keywords ($N_w$).  B, M,  and R-L represent BLEU, METEOR, and ROUGE-L, respectively.}
 \begin{tabular}{c|cccccc}
			\toprule[1pt] 
    $N_{w}$  & B-1 & B-2 & B-3 & B-4 & M & R-L \\ \hline
10 & 0.3793  & 0.2346  & 0.1540  & 0.1060  & 0.1625  & 0.2859  \\ \hline
15 & 0.3849  & 0.2265  & 0.1424  & 0.0937  & 0.1502  & 0.2668  \\ \hline
20 & 0.3769  & 0.2281  & 0.1478  & 0.1001  & 0.1519  & 0.2762  \\
			\bottomrule[1pt]
	\end{tabular}
	\label{tab:num_keywords}
\end{table}
\noindent\textbf{Effectiveness of keypatches.} In Table~\ref{tab:ablation}, when we further apply keypatches to the baseline model with keywords, the CXR generation performance boosts remarkably by about 0.5 in FID score, implying the effectiveness of keypatches provided by AdaMatch. To investigate the influence of the number of keypatches $N_p$, we analyze the CXR generation performance when the number of keypatches ranges from 5 to 15. As shown in Table~\ref{tab:num_keypatch}, AdaMatch-Cyclic achieves the best CXR generation performance with the FID score of 1.0916, when the number of keypatches is 5.

\begin{table}[t]
\small
	\renewcommand{\arraystretch}{1.2}
	\centering
	\caption{The analysis on the different number of keypatches($N_{p}$).}
 \begin{tabular}{c|c|c|c}
			\toprule[1pt] 
$N_{p}$	& 5	& 10	& 15\\ \hline
FID	& 1.0916 	& 1.6544 & 	1.6720 \\
			\bottomrule[1pt]
	\end{tabular}
	\label{tab:num_keypatch}
\end{table}

\begin{figure*}[tp]
  \centering
   \includegraphics[width=\linewidth]{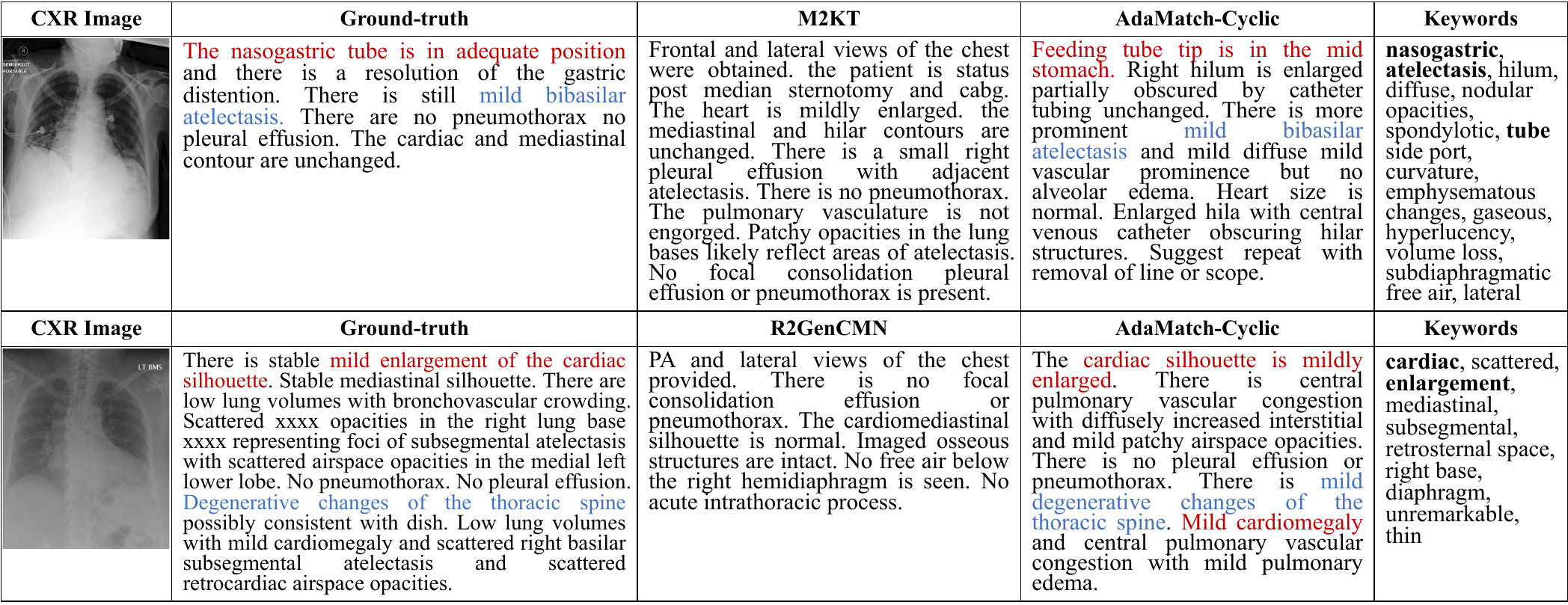}
   \caption{Qualitative comparison with existing methods in CXR-to-report generation on the MIMIC-CXR (1st row) and OpenI (2nd row) datasets. The texts in different colors show similar meanings. The keywords on the right are obtained from AdaMatch model.}
   \label{fig:cxr2report}
\end{figure*}
\begin{figure*}[tp]
  \centering
   \includegraphics[width=\linewidth]{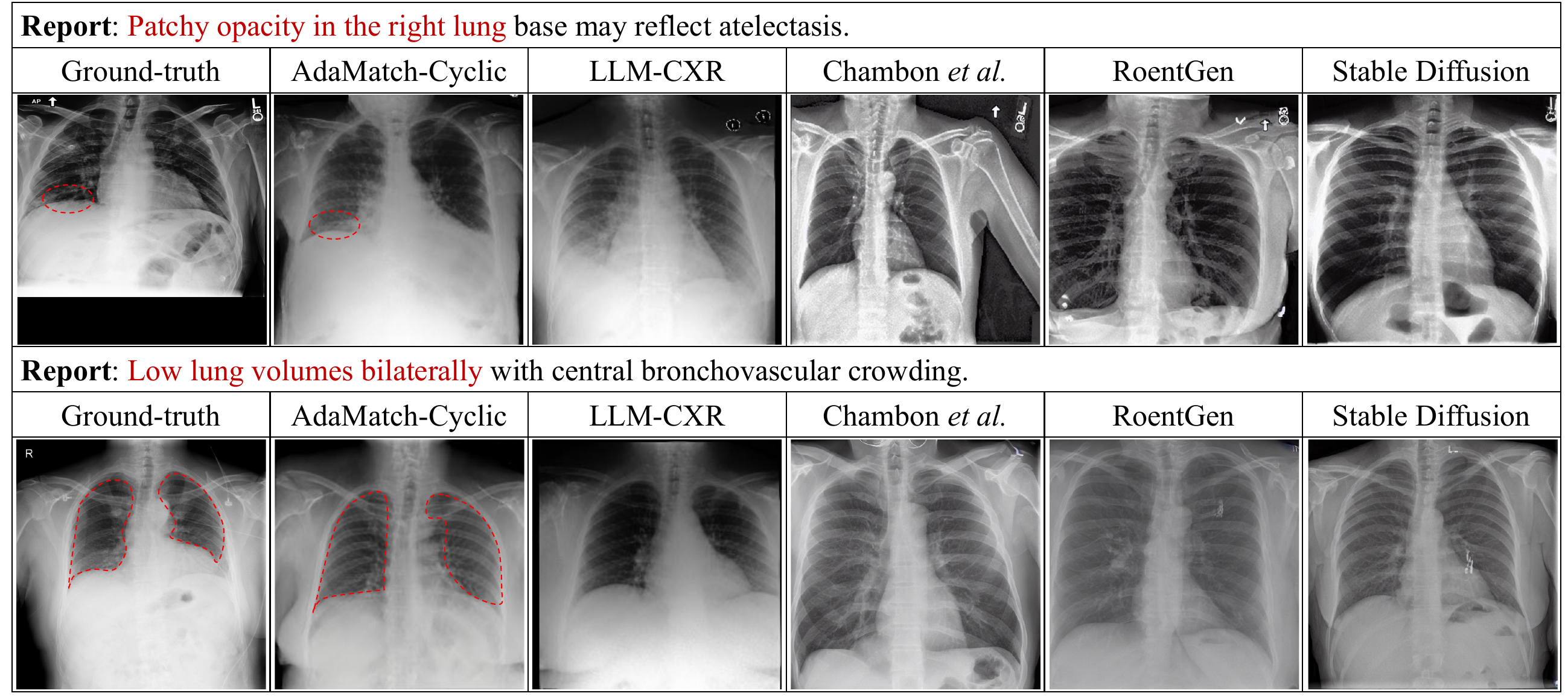}
   \caption{Visualization of the real and synthetic CXR images of the MIMIC-CXR (1st row) and the OpenI (2nd row) datasets.}
   \label{fig:report2cxr}
\end{figure*}

\begin{figure*}[tp]
  \centering
   \includegraphics[width=\linewidth]{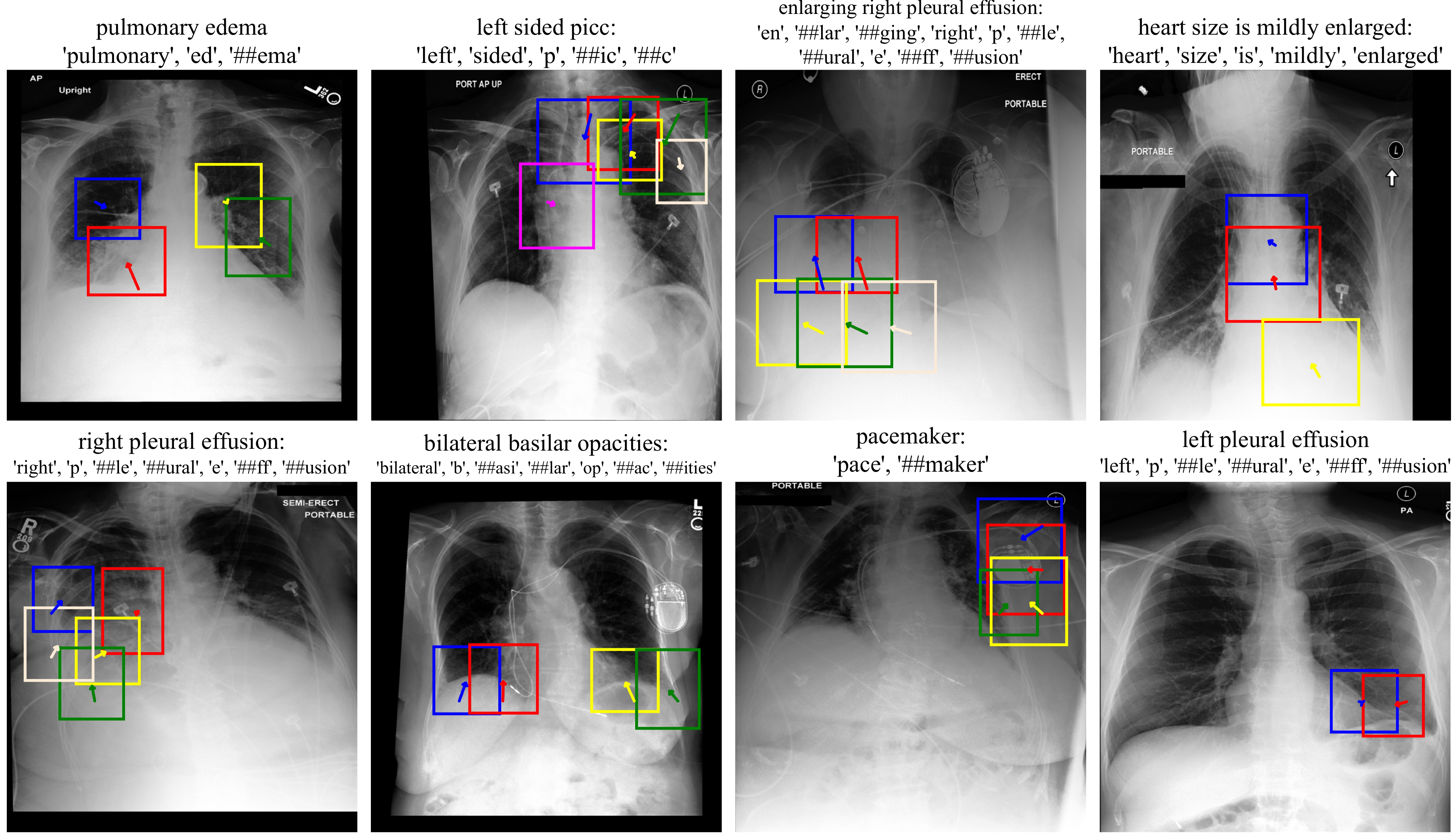}
   \caption{Visualization of texts and the corresponding adaptive patches. The boxes and arrows in different colors show adaptive patches and their center shifts from fixed patches.}
   \label{fig:adaptive}
\end{figure*}
\section{Visual Results}
\label{sec:vis}
To prove the effectiveness of AdaMatch-Cyclic for CXR-to-report and report-to-CXR generation tasks, we demonstrate some visual results for both tasks. Moreover, to further visualize the explanation provided by AdaMatch, i.e. the adaptive patches and the corresponding texts. \\
\noindent\textbf{CXR-to-Report Generation.}
As shown in Fig.~\ref{fig:cxr2report}, we compare CXR-to-report generation performance with existing methods on the MIMIC-CXR and OpenI datasets. In the first row, the proposed AdaMatch-Cyclic can capture the `Feeding tube tip' and `mild bibasilar atelectasis', while M2KT~\cite{yang2023radiology} cannot observe such device and lung abnormality. The second case of the OpenI dataset shows that `cardiac silhouette is mildly enlarged' and `degenerative changes of the thoracic spine' can be discovered by our method. These imply that our AdaMatch-Cyclic can generate a more comprehensive and complete medical report with the guidance of generated keywords in comparison to current methods.\\
\noindent\textbf{Report-to-CXR Generation.} 
Fig.~\ref{fig:report2cxr} visualizes the real and synthetic CXR images of the MIMIC-CXR and OpenI datasets in comparison with existing methods. As depicted in the first example, the CXR image generated by AdaMatch-Cyclic shows `patchy opacity in the right lung', while the CXR images generated by other methods do not include this pattern. In the second example, our AdaMatch-Cyclic can generate the CXR image with `low lung volumes'. These indicate the superiority of our AdaMatch-Cyclic over existing methods in report-to-CXR generation.
\\
\noindent\textbf{Adaptive Patches and Texts.}
In Fig.~\ref{fig:adaptive}, we visualize adaptive patches in AdaPatch, the textual words, and the textual tokens for different cases from the MIMIC-CXR datasets. We highlight the adaptive patches with bounding boxes in different colors. Each bounding box has an arrow inside to show the shift of center from the fixed patch to the adaptive patch. In the first example, adaptive patches cover the pulmonary edema. Meanwhile, adaptive patches of the second example show the correct position of PICC device in the left lung of the CXR image. These suggest that AdaMatch-Cyclic can show the correspondence between the adaptive patches and textual words to provide the correct explanation for the CXR-report generation.
\\

\section{Implementation Details}
\label{appendix:implement}
In the AdaMatch-Cyclic model, we first train the AdaMatch model and then use the frozen AdaMatch to train LLM. The AdaMatch model consists of an image encoder and a text encoder. We utilize the DPT-medium~\cite{chen2021dpt} as the image encoder that includes AdaPatch module in stage 2 and 3. The image encoder is pre-trained on the MIMIC-CXR dataset with the disease classification task. We adopt the pre-trained BioClinicalBERT~\cite{alsentzer2019publicly} as text encoder. The image and text encoders are followed by two convolutional layers with batch normalization and the ReLU activation function, respectively, to reduce the feature dimension to 256. The patch size $s$ is set to 4 for stage 1, and 2 for stage 2, 3, and 4. The number of sampled feature points $m$ is 3. We optimize the AdaMatch using LAMB optimizer ($\beta_1=0.9, \beta_2=0.999, \epsilon={10}^{-4}$). The cosine learning rate scheduler with the base learning rate ($lr\_b$) of $6\times10^{-3}$ is adopted to linearly warm up to the peak learning rate (i.e. $lr\_p = lr\_b \times \sqrt{\frac{bs\_all}{512}}$) during the first quarter of the total training epochs, where $bs\_all$ denotes the effective total batch size. The total training epoch is 15 and the per GPU batch size is 112. 
In AdaMatch-Cyclic, we use the pre-trained VQ-GAN~\cite{esser2021taming} model to encode CXR images into image tokens and decode image tokens into CXR images. We adopt the dolly-v2-3b~\cite{DatabricksBlog2023DollyV2} model as the pre-trained LLM. The LLM has 5,1845 token types with the first 5,0821 token types for text tokens and the rest 1,024 token types for image tokens. We add 5,0821 to each image token value encoded by VQ-GAN. To decode the image tokens into images, we subtract 5,0821 from the image tokens generated by LLM and feed image tokens to VQ-GAN decoder to obtain the generated CXR images. We train the LLM with the AdamW~\cite{loshchilov2018decoupled} optimizer. The learning rate is initialized as $5\times{10}^{-6}$ and the total training epoch is 5. The per GPU batch size is set to 24.
The hyper-parameters $\kappa_0, \kappa_1, \kappa_2, \kappa_3, \kappa_4$ are set as 200, 10, 1000, 20, and 5, respectively. 
In CXR-to-report generation, we extract keywords for each adaptive patch and use the keywords with the top 10 patch-word similarities as hints of LLM. In addition, we use 5 keypatches to guide the LLM in report-to-CXR generation.
All the experiments are conducted on 8 Nvidia A100 40GB GPUs.   There is no overlap of patients among different subsets. We discard the medical reports with fewer than 3 tokens from both datasets. All the CXR images with different sizes are resized to $256\times 256$ pixels. 



\end{document}